%% file: main.tex
\documentclass[acmtog]{acmart}

\AtBeginDocument{%
  \providecommand\BibTeX{{%
    \normalfont B\kern-0.5em{\scshape i\kern-0.25em b}\kern-0.8em\TeX}}}

\setcopyright{acmlicensed}
\acmJournal{TOG}
\acmYear{2024} \acmVolume{43} \acmNumber{4} \acmArticle{40} \acmMonth{7}\acmDOI{10.1145/3658222}

\usepackage{soul}
\usepackage{graphicx, animate}
\usepackage{caption}
\usepackage{subcaption}
\usepackage{multirow}
\usepackage{wrapfig}

\input{preamble}

\begin{document}

\title{EASI-Tex: Edge-Aware Mesh Texturing from Single Image}

\author{Sai Raj Kishore Perla}
\affiliation{%
 \institution{Simon Fraser University}
 \country{Canada}}
\email{srp7@sfu.ca}

\author{Yizhi Wang}
\affiliation{%
 \institution{Simon Fraser University}
 \country{Canada}}
\email{ywa439@sfu.ca}

\author{Ali Mahdavi-Amiri}
\affiliation{%
 \institution{Simon Fraser University}
 \country{Canada}}
\email{amahdavi@sfu.ca}

\author{Hao Zhang}
\affiliation{%
 \institution{Simon Fraser University}
 \country{Canada}}
\affiliation{%
 \institution{Amazon}
 \country{Canada}}
\email{haoz@sfu.ca}


\begin{abstract}
  \input{sections/0_abstract}
\end{abstract}



\begin{teaserfigure}
  \includegraphics[width=\textwidth]{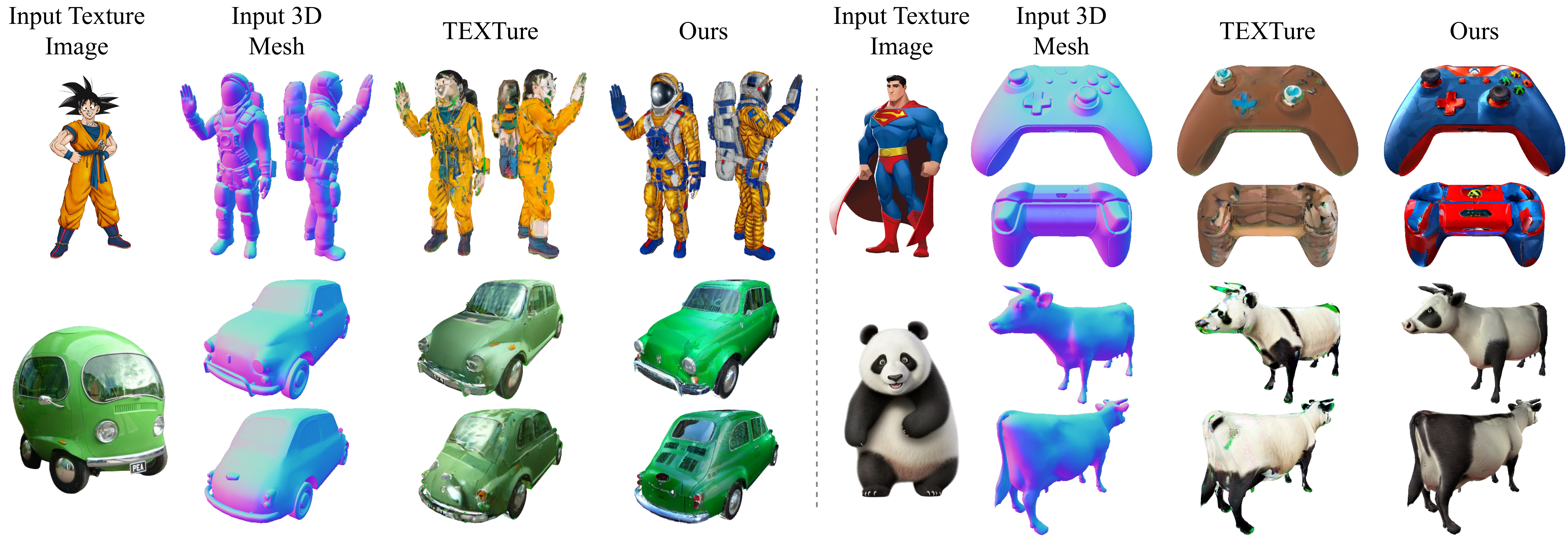}
  \caption{
  EASI-Tex seamlessly transfers an object's texture from a single RGB image to a given 3D mesh. Our method exhibits more natural textures for the 3D objects, respecting its semantics, and better preservation of geometric details and features, when compared to state-of-the-art alternatives such as TEXTure~\cite{Richardson2023Texture}. We show both within-category (left) and cross-category (right) texture transfers.
  \textit{Image Credits:} Goku~\cite{goku_img}, Superman~\cite{superman_img}, Green Car~\cite{green_car_img}, Panda~\cite{panda_img}.
  \textit{Mesh Credits:} Astronaut~\cite{astronaut_low_mesh}, Gaming Controller~\cite{xbox_mesh}, Car~\cite{compact_car_mesh}, Cow~\cite{DBLP:journals/tog/DeCarloFRS03}.
  }
  \label{fig:teaser}
\end{teaserfigure}


\setcopyright{acmlicensed}
\acmJournal{TOG}
\acmYear{2024} \acmVolume{43} \acmNumber{4} \acmArticle{40} \acmMonth{7}\acmDOI{10.1145/3658222}

\maketitle

\input{sections/1_introduction}

\input{sections/2_related_work}
\input{sections/3_method}

\input{sections/4_results}

\input{sections/5_conclusion}
\begin{acks}
\input{sections/6_acknowledgement}
\end{acks}

\bibliographystyle{ACM-Reference-Format}
\bibliography{main}


\end{document}

%% file: preamble.tex
\usepackage{color}

\definecolor{green}{rgb}{0, 0.5, 0}
\definecolor{orange}{rgb}{0.8, 0.6, 0.2}
\definecolor{red}{rgb}{1.0, 0.0, 0.0}
\definecolor{teal}{rgb}{0.0, 0.4, 0.4}
\definecolor{purple}{rgb}{0.65,0,0.65}
\definecolor{saffron}{rgb}{0.95,0.75,0.2}
\definecolor{turquoise}{rgb}{0.0,0.5,0.5}
\definecolor{black}{rgb}{0.0, 0.0, 0.0}
\definecolor{gray}{rgb}{0.5, 0.5, 0.5}

\newcommand{\rz}[1]{{\color{black}#1}}

%% file: sections/0_abstract.tex

We present a novel approach for \textit{single-image mesh texturing}, which employs a diffusion model with judicious conditioning to seamlessly transfer an object's texture from a single RGB image to a given 3D mesh object.
We do not assume that the two objects belong to the same category, and even if they do, there can be significant discrepancies in their geometry and part proportions.
Our method aims to rectify the discrepancies by conditioning a pre-trained Stable Diffusion generator with edges describing the mesh through ControlNet, and features extracted from the input image using IP-Adapter to generate textures that 
respect the underlying geometry of the mesh and the input texture without any optimization or training.
We also introduce {\em Image Inversion\/}, a novel technique to quickly personalize the diffusion model for a single concept using a \textit{single image}, for cases where the pre-trained IP-Adapter falls short in capturing all the details from the input image faithfully.
Experimental results demonstrate the efficiency and effectiveness of our edge-aware single-image mesh texturing approach, coined EASI-Tex, in preserving the details of the input texture on diverse 3D objects, while respecting their geometry.

\noindent
\textbf{Project Page:} \href{https://sairajk.github.io/easi-tex/}{\texttt{https://sairajk.github.io/easi-tex}}




%% file: sections/1_introduction.tex
\section{Introduction}
\label{sec:intro}

With the rapid advances in generative AI, the demand for high-quality textured 3D shapes has intensified in recent years. 
In particular, the wave of developments in this problem domain have been propelled by the emergence of large language and vision-language models~\cite{du2022survey, zhang2023visionlanguage}, as well as the unparalleled quality of results obtained by powerful generators such as diffusion models~\cite{rombach2022high, yang2023diffusion}. 
Lately, the most popular line of approaches~\cite{Chen_2023_ICCV, poole2022dreamfusion, tang2023makeit3d, wang2023prolificdreamer} resort to text-guided image diffusion models to optimize a neural radiance field (NeRF)~\cite{mildenhall2020nerf} through differentiable rendering.
However, the resulting geometries and textures are often low-resolution, blurry, and not feature-preserving. 
They are also hardly reusable for downstream tasks, since NeRF~\cite{mildenhall2020nerf}, and similarly, 3D Gaussian splats~\cite{kerbl2023gs}, are both rendering primitives, not modeling primitives. 
In addition, the generative processes themselves lack fine-grained control, especially with text prompting, which can be ambiguous.

In this paper, we study the problem of {\em single-image mesh texturing\/} as a means to endow high-quality 3D mesh models with rich textures 
transferred from an abundant supply of online images. 
Specifically, we take as input a polygonal mesh object $\mathbf{M}$ and a single image $\mathbf{I_{tex}}$ of a different object, captured from an {\em arbitrary view\/}.
The output is a fully textured version of $\mathbf{M}$ as guided by $\mathbf{I_{tex}}$, without any geometric changes to $\mathbf{M}$. 
Clearly, for the texture transfer to be a reasonable one, there ought to be a meaningful correspondence between the object in the input image and $\mathbf{M}$. 
Yet, we do not assume that the two objects belong to the same category, and even if they do, there can be significant discrepancies in their geometry and part proportions, as shown in Fig.~\ref{fig:teaser}. 
Our method aims to not only rectify the discrepancies to ensure plausible and natural texture transfer, while faithful to $\mathbf{I_{tex}}$, but also respect the geometry of $\mathbf{M}$ to produce clean and sharp textures over the 3D model.

We treat our texturing problem as a {\em generative\/} task since the input image only contains partial information from a single view, and even implausible
textures for the input 3D model due to potential mismatching. 
To this end, we develop a novel approach employing a diffusion model with judicious conditioning~\cite{ye2023ip, zhang2023adding} to achieve high-quality texture transfers which respect the input texture, as well as the geometry and semantics of the input 3D model.
Specifically, 
\begin{itemize}
\item Leveraging a pre-trained Stable Diffusion (SD)~\cite{rombach2022high} generator, our method is capable of transferring textures to a given mesh even in the absence of a direct guide from the single-view input image.
\item To preserve the \textit{identity} of the 3D model, we extract edges characterizing the geometry of the input mesh and use them as a conditioning signal to the diffusion model through ControlNet~\cite{zhang2023adding} while texturing.
\item We further condition the generation process with both features extracted from the input image and from a descriptive text prompt using IP-Adapter~\cite{ye2023ip}. Such a text compatible image prompt adapter allows us to use a single image as prompt, without any optimization or training, to efficiently guide the diffusion model while texturing.
\end{itemize}

Our edge-aware single-image mesh texturing approach, coined EASI-Tex, is an efficient and optimization-free alternative to personalization~\cite{gal2022textual, ruiz2023dreambooth} based texture transfer~\cite{Richardson2023Texture}, while preserving the geometric and semantic \textit{identity} of the given input mesh.
The latter is accomplished through edge conditioning, rather than depth conditioning~\cite{chen2023text2tex, Richardson2023Texture}, which is weaker in terms of \textit{identity} preservation and respecting the fine geometric details from the mesh. 
Indeed, most high-quality meshes are created by artists by keeping the geometric structure and semantic features of the object in mind. 
Our edge-aware texturing approach is not only designed to harness these geometric and semantic attributes, but also opens doors to new applications such as ``semi-retexturing'', where the edges presented in an input {\em textured\/} mesh can serve as a conditioning signal.
We also introduce \textit{Image Inversion}, a novel personalization technique to quickly adapt the diffusion model for a single concept using a \textit{single-image}, for cases where the pre-trained IP-Adapter~\cite{ye2023ip} falls short in capturing all the details from the input texture image faithfully.

Experimental results reveal that existing
methods~\cite{Richardson2023Texture} struggle to transfer textures from a single-image accurately or remain faithful to the 3D model's geometry.
We solve both these problems and demonstrate the effectiveness of our approach in preserving the details of the input texture image, $\mathbf{I_{tex}}$, on diverse 3D objects.
We evaluate the performance of our approach through both qualitative and quantitative comparisons, and illustrate the effect of various important components in our method through ablation studies.
Our results (see Sec.~\ref{sec:results}), generated by judicious conditioning of diffusion models, are more color accurate, consistent across different viewpoints, preserve the input texture, and respect the underlying geometry of the input 3D mesh better than the baselines.
Our proposed approach is also the most efficient when it comes to transferring textures, both in terms of time and the number of input images required.

%% file: sections/2_related_work.tex
\section{Related Work}
\label{sec:related}

Our work is situated in the context of several relevant prior research. We begin by giving an overview of methods related to texture synthesis and generation. Next, we explore techniques that generate 3D shapes and textures simultaneously. Finally, we investigate works that specifically tackle the challenge of transferring or generating textures for a given 3D mesh as input.

\subsubsection*{Texture Synthesis}
There exist numerous techniques that generate 2D textures~\cite{8659343, de1997multiresolution, efros1999texture, heeger1995pyramid, zhu1998filters}, or geometric textures for 3D objects~\cite{hertz2020deep}. 
To synthesize 2D textures for 3D shapes, from a given textured mesh as reference, some methods~\cite{Chen:2012:NTT, lu2007context, mertens2006texture} leverage the correlation between geometric features and texture values of the reference mesh.
However, these techniques are only effective for simple textures and fail to transfer complex, multicolored textures onto 3D shapes.

\subsubsection*{Textured Mesh Generation}
The prevalent trend in current research revolves around text-conditioned image and 3D shape generation.
DreamFusion~\cite{poole2022dreamfusion} employs a pre-trained text-to-image diffusion model to propose Score Distillation Sampling (SDS) and train a NeRF~\cite{mildenhall2020nerf} model for 3D shape generation. Magic3D~\cite{lin2023magic3d} extends this approach to 3D meshes by incorporating neural marching tetrahedra~\cite{shen2021deep}.
Fantasia3D~\cite{Chen_2023_ICCV} employs a disentangled approach, separating the modeling and learning of geometry and appearance, for high-quality text-to-3D content creation.
However, decoupling appearance learning also enables texture generation for given 3D shapes as input.
ProlificDreamer~\cite{wang2023prolificdreamer} introduces Variational Score Distillation (VSD), while also utilizing a disentangled approach for 3D shape generation. It significantly improves the distillation of diffusion models, enabling the generation of photo-realistic textures with rich structure and complex effects.
%
To avoid encoding images to the latent space while training NeRF using a latent diffusion model~\cite{rombach2022high}, Latent-NeRF~\cite{metzer2023latent} takes a novel approach by training a NeRF directly in the latent space of the diffusion model.
It also proposes to generate textures on a given 3D shape, through \textit{Latent-Paint}, by optimizing a latent texture image for the 3D shape instead of a NeRF.

\subsubsection*{3D Mesh Texturing}
There have been several works on texture transfer and generation for a given 3D shape as input~\cite{cao2023texfusion, chen2022AUVNET, photoshape2018}.
While earlier works~\cite{bokhovkin2023mesh2tex, siddiqui2022texturify} relied on an adversarial framework (such as GANs~\cite{goodfellow2014generative}), more recent techniques~\cite{chen2023text2tex, Richardson2023Texture} employ pre-trained image diffusion models for this purpose.

Texture Fields~\cite{oechsle2019texture} 
is one of the early works on texture transfer, they 
represent texture as a continuous neural field and employ an adversarial strategy~\cite{goodfellow2014generative} to train their network. However, the generated textures are often smooth and lack high-frequency details.
Mesh2Tex~\cite{bokhovkin2023mesh2tex} also uses a GAN-based architecture for generating textures on a given 3D shape. It further demonstrates the ability to transfer textures from an image through test-time optimization, leveraging global and patch-level style losses. However, all GAN-based methods are usually trained from scratch on small-scale datasets which often limits the overall quality of the generated textures, and inhibits them from generalizing well for cross-category texture transfers.

More recent works harness the capabilities of pre-trained image diffusion models to transfer and generate textures on a given 3D mesh. 
TEXTure~\cite{Richardson2023Texture} and Text2Tex~\cite{chen2023text2tex} are two recent optimization-free 3D mesh texturing techniques which follow a very similar texturing strategy. 
They both generate textured views of the mesh using a diffusion model and \textit{paste} them onto the 3D shape for texturing. 
However, they do so iteratively, all around the mesh with overlapping viewpoints, and maintain a custom \textit{keep-refine-generate} mask to produce consistent textures on the 3D shape.

While TEXTure also facilitates texture transfer from images through 
personalization~\cite{gal2022textual, ruiz2023dreambooth},
it relies on multiple images, specifically multi-view images of an object, which may not always be accessible.
Moreover, 
it 
requires optimizing the network on each set of images intended for texture transfer, resulting in an additional overhead of time, compute, and memory.
To address these challenges, we propose leveraging a pre-trained IP-Adapter~\cite{ye2023ip} to guide the diffusion model while texturing. 
This allows us to directly use a single image as prompt, without the need for any optimization or training.

Furthermore, existing works on 3D shape texturing using diffusion models either exclusively rely on text prompts~\cite{Chen_2023_ICCV, wang2023prolificdreamer, metzer2023latent} or additionally employ some spatial information, like depth~\cite{Chen_2023_ICCV, Richardson2023Texture}, as a conditioning signal.
Although incorporating depth improves the shape-texture consistency 
, we argue depth to be a weak conditioning signal as it is usually smooth and lacks the fine details from the mesh (see Fig.~\ref{fig:depth_vs_edge}).
This results in visually pleasing but inconsistent textures that may not respect or semantically correspond to the underlying geometry of the input mesh, often changing its \textit{identity} (see Fig.~\ref{fig:clip_similarity}).
To tackle this, we propose to use edges, a better mesh descriptor, to generate textures that are more faithful to the geometry and semantics of the input mesh.

%% file: sections/3_method.tex
\section{Method}
\label{sec:method}

\input{figures/pipeline}

Given an untextured 3D mesh, $\mathbf{M}$, of an object and an exemplar texture image, $\mathbf{I_{tex}}$, as input, our goal is to transfer the texture from $\mathbf{I_{tex}}$ to $\mathbf{M}$ while respecting the semantic parts and geometric features of the 3D object. The process of transferring texture involves two main steps: (1) understanding the content in $\mathbf{I_{tex}}$ and extracting the texture information accurately, and (2) understanding the semantics and geometry of $\mathbf{M}$, and transferring the extracted texture faithfully to the corresponding parts. The challenge arises when the content in $\mathbf{I_{tex}}$ and $\mathbf{M}$ are different, requiring the texturing pipeline to balance between texture transfer and generation to produce plausible textures.
We believe this can be achieved by enhancing the model's understanding of the underlying content, $\mathbf{I_{tex}}$ and $\mathbf{M}$, during the texturing process.

We aim to solve this problem using better conditioning signals as input to the diffusion model. 
Specifically, we diverge from the commonly used conditioning signals, such as text prompts \cite{Chen_2023_ICCV, wang2023prolificdreamer} or text prompts with depth maps \cite{chen2023text2tex, Richardson2023Texture}, to use \textit{stronger} conditioning signals -- the input texture image ($\mathbf{I_{tex}}$), edges extracted from the 3D mesh ($\mathbf{M}$), and text prompt -- directly as input to the diffusion model.
This improves the quality of texture transfer  significantly, and also allows us to respect the semantics and geometry of the 3D mesh better.
We further introduce \textit{Image Inversion}, a novel personalization technique using a \textit{single-image} for cases where the pre-trained models fall short in capturing all the details from the input texture image faithfully.
In the subsequent sections, we first provide the fundamental concepts and tools relevant to this work, followed by the details of image and edge conditioning, along with the technique of \emph{Image Inversion} for single-image guided 3D mesh texturing.

\subsection{Preliminary}
\label{subsec:preliminary}

\subsubsection{Diffusion Models:}
\label{subsubsec:diffusion_models}

Diffusion models~\cite{dhariwal2021diffusion, ho2020denoising} operate in a two-step process, a \textit{forward} process adds noise to a clean image, while the \textit{backward} process takes this noisy image as input to estimate the added noise using a neural network.
Stable Diffusion (SD)~\cite{rombach2022high} is a \textit{latent diffusion model} where the diffusion noising and denoising takes place in the \textit{latent} space. It employs a pre-trained \textit{variational autoencoder} (VAE)~\cite{kingma2014autoencoding} to first encode a clean image, $x_0$, to its latent representation, $z_0$.
This latent is noised to $z_t = \sqrt{\bar{\alpha}_t}z_0 + (\sqrt{1 - \bar{\alpha}_t})\epsilon$ as per some random time step $t \sim \mathbf{U}(0, T)$, where $\bar{\alpha}_t$ is computed from a predefined noise schedule~\cite{ho2020denoising}, $\epsilon \sim \mathcal{N}(0, \mathbf{I})$ is the Gaussian noise, and $T$ is the total number of noising steps in the \textit{forward} process.
The noised latent, $z_t$, is then passed through a modified U-Net~\cite{ronneberger2015u}, $\mathcal{U}(\cdot)$, to predict the noise added to the clean input, $z_0$.
SD is also conditioned on an input text prompt, $p$, to generate guided outputs. The text prompt, $p$, is first tokenized and passed through a CLIP~\cite{radford2021learning} text encoder, $\mathcal{C}_{txt}(\cdot)$, to generate $\mathbf{f_{txt}} = \mathcal{C}_{txt}(\texttt{tokenize}(p))$. 
These text prompt features, $\mathbf{f_{txt}}$, are then incorporated into SD's U-Net, $\mathcal{U}(\cdot)$, using cross-attention~\cite{vaswani2017attention} layers as:
\begin{equation}
    \begin{gathered}
        \label{eqn:sd_cross_attn}
        Q=W_Q z_t; \quad K=W_K \mathbf{f_{txt}}; \quad V=W_V \mathbf{f_{txt}} \\
        Z = \mathtt{Attention}(Q, K, V) = \mathtt{Softmax}\left(\textstyle\frac{QK^T}{\sqrt{d}}\right)\cdot V
    \end{gathered}
\end{equation}
where $W_Q$, $W_K$, and $W_V$ are the learnable weights for projection, $d$ is the dimension of each feature vector in $\mathbf{f_{txt}}$, and $Z$ is the output of the cross-attention operation.
Given these, SD is trained using the loss function:
\begin{equation}
    \begin{gathered}
        \label{eqn:sd_loss}
        \mathcal{L}_{SD} = \mathbb{E}_{z_t, t, \epsilon, p} \left[ {\lVert \mathcal{U}(z_t, t, \mathbf{f_{txt}}) - \epsilon \rVert}_2^2 \right]
    \end{gathered}
\end{equation}
During inference, SD is fed with Gaussian noise which is iteratively denoised to generate an image conditioned on the input text prompt.

\subsubsection{IP-Adapter:}
\label{subsubsec:ip_adapter}

IP-Adapter~\cite{ye2023ip} introduces an efficient image conditioning adapter for pre-trained text-to-image diffusion models, supplementing conventional text conditioning. 
It employs a decoupled cross-attention mechanism for text and images, where for each cross-attention layer used to incorporate the text prompt features, $\mathbf{f_{txt}}$, into SD's U-Net, an additional cross-attention layer is appended alongside to incorporate features from the input image in a similar manner. 
The IP-Adapter consists of three main components: $(1)$ a pre-trained CLIP~\cite{radford2021learning} image encoder, $\mathcal{C}_{img}(\cdot)$,  to extract features from the input image, $\mathbf{I_{tex}}$ in our case, $(2)$ a projection network, $\mathcal{P}_{ip}(\cdot)$, to decompose these features into a sequence of image tokens, $\mathbf{f_{tex}} = \mathcal{P}_{ip}(\mathcal{C}_{img}(\mathbf{I_{tex}}))$, and $(3)$ the newly added cross-attention~\cite{vaswani2017attention} layers to incorporate the image tokens, $\mathbf{f_{tex}}$, into SD's U-Net, $\mathcal{U}(\cdot)$, as:
\begin{equation}
    \begin{gathered}
        \label{eqn:ip_cross_attn}
        Q =W_Q z_t; \quad K'=W'_K \mathbf{f_{tex}}; \quad V'=W'_V \mathbf{f_{tex}} \\
        Z_{new} = \mathtt{Attention}(Q, K, V) + \lambda_{ip} \mathtt{Attention}(Q, K', V')
    \end{gathered}
\end{equation}
where $W_Q$, $W'_K$, and $W'_V$ are the learnable weights for projection, and $Z_{new}$ is the updated output of the cross-attention operations. 
It is worth noting that a common $Q$ serves both text and image cross-attention layers, and $\lambda_{ip}$ is a hyperparameter that controls the influence of the input image on the diffusion output, it is set to $1$ while training and can be adjusted as required during inference.
Lastly, while IP-Adapter shows experiments only with images, in this work, we extend it to 3D and showcase its ability to transfer complex textures between disparate objects in a consistent manner.

\subsubsection{ControlNet:}
\label{subsubsec:controlnet}

ControlNet~\cite{zhang2023adding} introduces a neural network framework to incorporate spatial conditioning controls, such as edges, normals, or depth, to pre-trained image diffusion models, like SD. It maintains the quality and capabilities of the pre-trained diffusion model by locking their parameters and creating a trainable copy of the encoding layers, the \textit{ControlNet}. The spatial conditioning signal is encoded by the ControlNet and transmitted to the pre-trained diffusion model through residual connections as: 
\begin{equation}
    \begin{gathered}
        F'_{un} = F_{un} + \lambda_{cn}F_{cn}
    \end{gathered}
\end{equation}
where $F_{un}$ are the features from SD's U-Net, $F_{cn}$ are the features from the ControlNet, and $F'_{un}$ are the updated U-Net features. $\lambda_{cn}$ is a hyperparameter that regulates the influence of the conditioning signal on the pre-trained diffusion model, it is set to $1$ while training and can be adjusted as required during inference.

Lastly, both IP-Adapter and ControlNet are based on SD v1.5, and while trained independently, both are trained by freezing the parameters of the pre-trained diffusion model.
This allows these models to generalize not only to other custom models fine-tuned from the same base diffusion model (\textit{i.e.} SD v1.5), but also to controllable generation using other controllable tools, such as each other, allowing for an easy combination of image prompting, using IP-Adapter, with structure control, through ControlNet.

\subsection{Edge Conditioning}
\label{subsec:edge_cond}

In this work, we propose to adopt a robust conditioning signal — edges — for texturing.
Compared to depth, edges, as we propose to compute them, can describe the details of an untextured 3D mesh better and also serve as a \textit{stronger} conditioning signal to the diffusion model. 
This allows us to generate textures that are clean, sharp, and align more closely with the geometry and semantics of the underlying 3D model.
As illustrated in Fig.~\ref{fig:depth_vs_edge}, the depth map is usually smooth and many of the details, like the watch, belt, and other accessories of the astronaut, that are clearly visible with edges can be barely seen in the depth map, and the same is reflected in the corresponding textured views as well.

\input{figures/depth_vs_edgemap}

We harness various geometric features, like the connected components (CCs), depth, and normals, of the mesh to extract rich edge maps for conditioning.
To extract edges from CCs, we first identify all the CCs of the mesh and assign a random color to each CC. This mesh with randomly colored CCs is then rendered from a desired viewpoint to generate an RGB image, which is passed through a Canny edge detector to extract the edges. However, during such a random assignment of colors, neighbouring CCs may get similar or even the same color which hinders the extraction of edges between them. To tackle this, we repeat the above process multiple times with different random color assignments to the CCs, and take a union of all the edges extracted from these iterations. This ensures we capture all the edges between CCs faithfully.
To extract edges from depth and normals, we first render the depth/normals of the mesh from the desired viewpoint and normalize the values to $[0, 255]$. This normalized depth/normal map is then passed to a Canny edge detector to extract the edges.
The final edge map, $\mathbf{I_{edge}}$, is computed as the union of all the edges extracted from various geometric attributes (CCs, depth, and normals) of the input mesh.

As shown in Fig.~\ref{fig:pipeline}, the final edge map, $\mathbf{I_{edge}}$, is input to a pre-trained Canny ControlNet~\cite{zhang2023adding} to produce edge-conditioning features, $\mathbf{f_{geo}}$, which are fed to the SD model to generate the textured view.
While one could argue to incorporate each geometric attribute (CCs, depth, and normals) separately, using multiple ControlNets, there are three main drawbacks associated with such an approach: (1) a pre-trained ControlNet might not be available for every geometric attribute we want to use; for instance, there is no pre-trained ControlNet model available for CCs, (2) using multiple ControlNets makes the forward pass, and hence the texturing process, slower, (3) using multiple ControlNets with different inputs leads to a set of competing models trying to impose their output at the same region, resulting in a degraded final outcome.
In contrast, our approach leverages information from multiple geometric features efficiently, using a \textit{single} ControlNet model. 
Our method is also very flexible, allowing for an easy inclusion or exclusion of geometric attributes while computing edges.


\subsection{Image Conditioning \& Image Inversion}
\label{subsec:img_cond}

To condition the diffusion model on the input texture image, $\mathbf{I_{tex}}$, we propose to employ IP-Adapter~\cite{ye2023ip} which allows us to use a single image as input prompt, without any additional training or optimization. 
IP-Adapter first extracts the image tokens, $\mathbf{f_{tex}} = \mathcal{P}_{ip}(\mathcal{C}_{img}(\mathbf{I_{tex}}))$, which are incorporated into SD's U-Net using cross-attention layers as described in Sec.~\ref{subsubsec:ip_adapter}.

\subsubsection{Image Inversion:}
\label{subsubsec:img_inv}
We found the pre-trained IP-Adapter~\cite{ye2023ip} to occasionally miss out on fine details while capturing textures, especially when the input image has small non-recurring designs or patterns.
One of our main contributions is to mitigate this issue by introducing \textit{Image Inversion}, which is a novel 
approach to quickly personalize a pre-trained SD model with IP-Adapter for a single concept using a single image. While Textual Inversion involves learning text token(s) directly~\cite{gal2022textual, voynov2023P+} or learning a network that generates the text tokens~\cite{alaluf2023neural}, the same is not possible with IP-Adapter. To tackle this, we propose to fine-tune the projection network, $\mathcal{P}_{ip}(\cdot)$, of the IP-Adapter and SD's U-Net, $\mathcal{U}(\cdot)$, using the single texture image, $\mathbf{I_{tex}}$, for personalization.
The pipeline for \textit{Image Inversion} is shown in Fig.~\ref{fig:pipeline}, and the optimization objective, $\mathcal{L}_{II}$, is given below with the trainable networks shown in \textcolor{red}{red}:

\begin{equation}
    \begin{gathered}
        \label{eqn:img_inv_loss}
        \mathbf{f_{tex}} = \textcolor{red}{\mathcal{P}_{ip}(}\mathcal{C}_{img}(\mathbf{I_{tex}})\textcolor{red}{)} \\
        \mathcal{L}_{II} = \ \mathbb{E}_{z_t, t, \epsilon, p, \mathbf{I_{tex}}} \left[ {\lVert \textcolor{red}{\mathcal{U}(} z_t, t, \mathbf{f_{txt}}, \mathbf{f_{tex}} \textcolor{red}{)} - \epsilon \rVert}_2^2 \right]
    \end{gathered}
\end{equation}

To prevent overfitting, we use structural augmentations such as random flip, random resize, and random rotation around the center by a small degree for the texture image during personalization. 
Following \cite{gal2022textual}, we randomly select prompts from a set of neutral text prompts describing the 
input image while fine-tuning.
Once fine-tuned, we freeze both the projection network, $\mathcal{P}_{ip}(\cdot)$, and SD's U-Net, $\mathcal{U}(\cdot)$, which are then used for texturing meshes as described in Sec.~\ref{subsec:tex_mesh}.
It is worth nothing that \textit{Image Inversion} is fairly generalizable and can be used to personalize the network for a single concept using multiple images as well.

\subsection{Texturing Meshes}
\label{subsec:tex_mesh}

After acquiring the conditioning signals, $\mathbf{f_{geo}}$ and $\mathbf{f_{tex}}$, from ControlNet~\cite{zhang2023adding} and IP-Adapter~\cite{ye2023ip}, we employ them to generate textured views of the 3D shape using pre-trained or personalized networks, as shown in Fig.~\ref{fig:pipeline}.
We use Text2Tex~\cite{chen2023text2tex} for texturing, which iteratively generates and \textit{pastes} textured views onto the input 3D mesh. Notably, this process doesn’t require any optimization and is very fast.
One can optionally use optimization-based techniques~\cite{Chen_2023_ICCV, wang2023prolificdreamer} for texturing, to further enhance the details.
However, such methods can take hours to generate textures while optimization-free approaches, like Text2Tex, only need a few minutes.

%% file: figures/pipeline.tex
\begin{figure*}
    \centering

    \includegraphics[width=\textwidth, keepaspectratio]{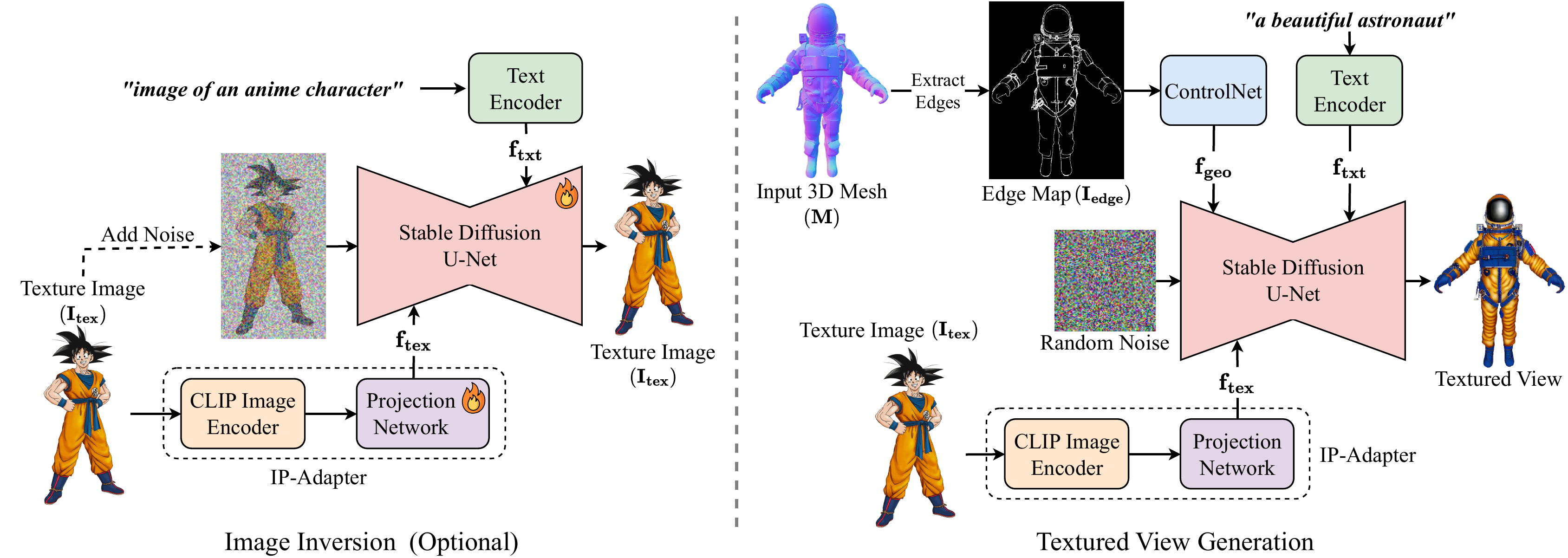}

    \caption{
    The pipeline of EASI-Tex for \textit{Image Inversion} (left) and generating textured views (right).
    \textit{Image Inversion} is an optional step that involves fine-tuning parts of our network: Stable Diffusion's U-Net and IP-Adapter's Projection Network (indicated by \includegraphics[width=7pt, keepaspectratio]{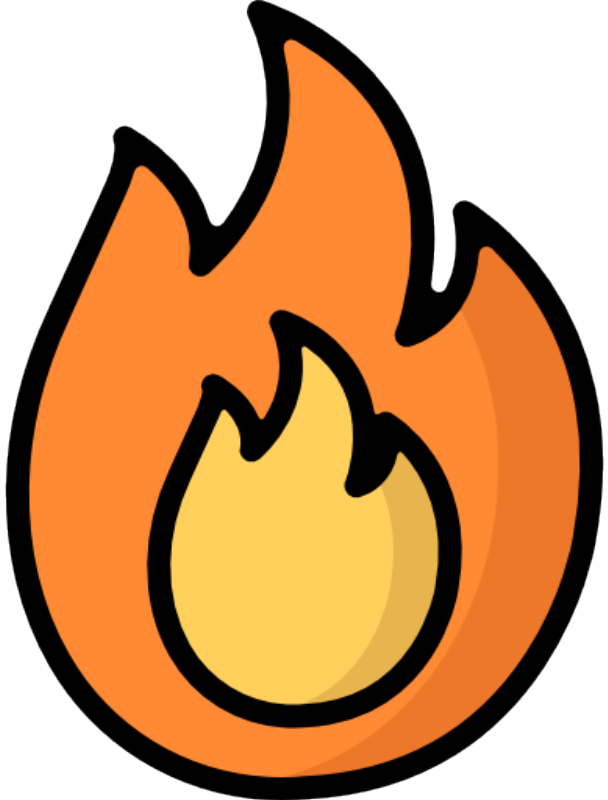}). 
    It uses a single image ($\mathbf{I_{tex}}$) to quickly personalize the pre-trained networks for the input texture.
    The texture generation network
    takes an untextured 3D mesh ($\mathbf{M}$), a reference texture image ($\mathbf{I_{tex}}$), and a descriptive text prompt for $\mathbf{M}$ as input to generate a textured view of the mesh as output.
    Stable Diffusion's U-Net is guided by three conditioning signals: (1) the edge map of a sampled view of the 3D mesh, through $\mathbf{f_{geo}}$, (2) the input texture image, through $\mathbf{f_{tex}}$, and (3) a text prompt describing the 3D mesh, through $\mathbf{f_{txt}}$.
    \textit{Image Credits:} Goku~\cite{goku_img}, Fire~\cite{fire_img}.
    \textit{Mesh Credits:} Astronaut~\cite{astronaut_high_mesh}.
    }
    \label{fig:pipeline}
\end{figure*}


%% file: figures/depth_vs_edgemap.tex
\begin{figure}
    \centering

    \includegraphics[width=0.8\columnwidth, keepaspectratio]{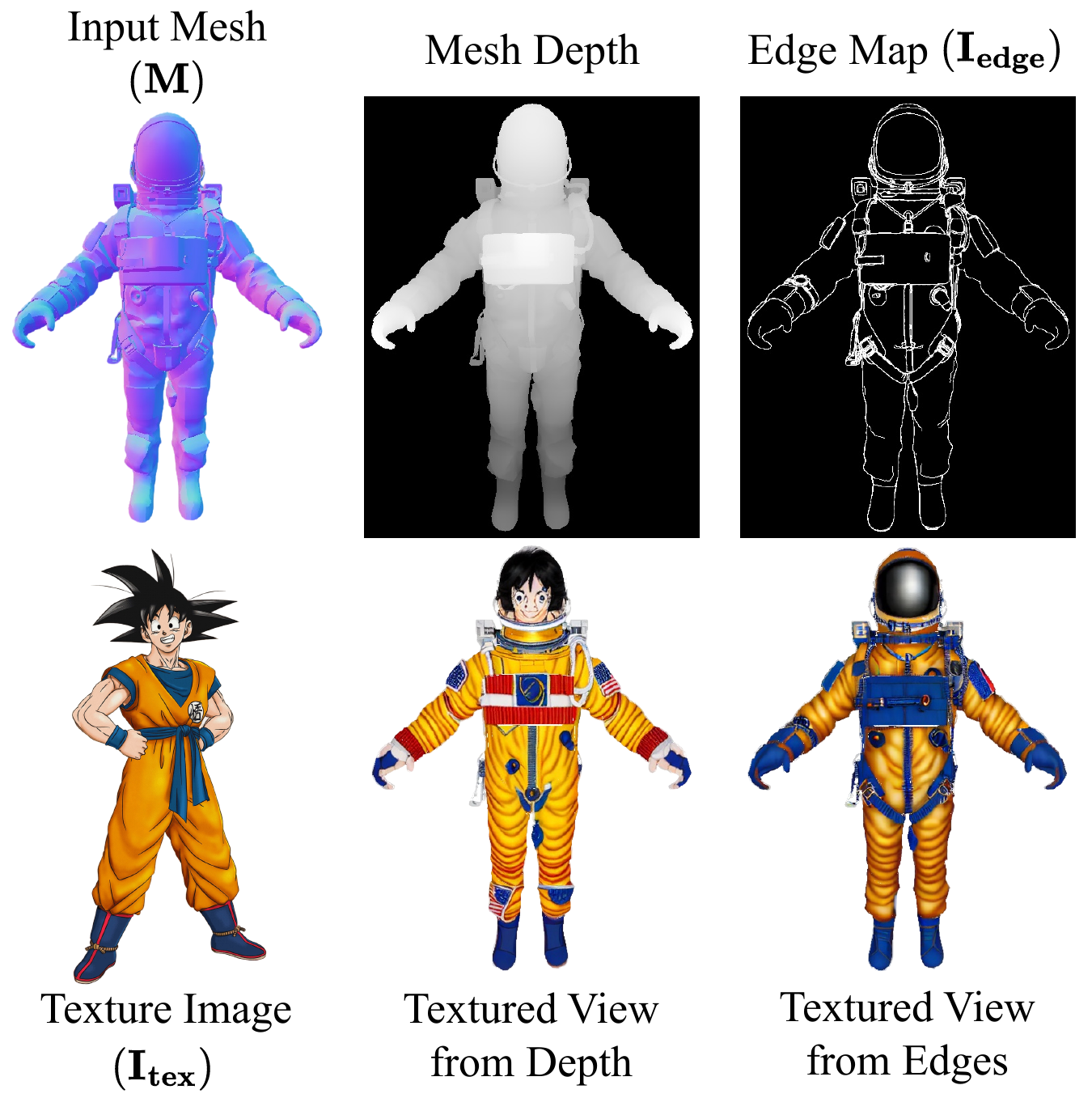}

    \caption{
    The depth and edge map extracted from an untextured 3D mesh ($\mathbf{M}$) of an astronaut, and their corresponding textured views for a given texture image $\mathbf{I_{tex}}$.
    It is evident that when edges are used, the resulting texturization is more faithful to the underlying geometry of the 3D model (\textit{e.g.}, head, belt straps, watch, \textit{etc.}).
    \textit{Image Credits:} Goku~\cite{goku_img}.
    \textit{Mesh Credits:} Astronaut~\cite{astronaut_high_mesh}.
    }
    \label{fig:depth_vs_edge}
\end{figure}

%% file: sections/4_results.tex
\input{figures/visual_comp}
\section{Results and Evaluation}
\label{sec:results}

\subsection{Datasets}
\label{subsec:datasets}
We perform experiments on diverse 3D objects primarily chosen from the Objaverse~\cite{objaverse, objaverseXL}, along with some other sources~\cite{DBLP:journals/tog/DeCarloFRS03, DBLP:conf/siggraph/PraunFH00}.
To assess the generality and capability of our approach, we curate a mixed set of 3D models for evaluation, including both organic (astronauts, animals) and human-made (gaming controllers, vehicles, chairs, etc.) objects.
We also include detailed (\textit{e.g.}, astronauts) as well as simpler water-tight meshes with a single connected component (\textit{e.g.}, cow), recognizing the challenge in detecting edges in simpler models compared to the more detailed ones.
The texture images are randomly sourced from the internet, previous works~\cite{gal2022textual, yang2015large}, and renderings of 3D models from the Objaverse \cite{objaverse, objaverseXL} dataset.
We mask the background of these images with a constant color, usually white, to prevent unintended noise during \textit{Image Inversion} and texturing.

The images and 3D shapes are also paired randomly, for in-category texture transfer, images containing objects similar to the 3D model’s category are selected. However, for cross-category texture transfer, no particular rules apply.
In fact, we demonstrate the possibility of transferring textures between disparate image-shape pairs that exhibit significant differences in geometry and semantic parts, for instance, from an image of Superman to a 3D model of a gaming controller (see Fig.~\ref{fig:teaser}).
This illustrates the ability of our method to generalize well and tackle challenging image-shape pairs.

\subsection{Evaluation}
\label{subsec:evaluation}

\input{figures/clip_similarity}

It is non-trivial to evaluate texture transfer from a single image to a 3D mesh due to the differences in geometry, viewpoint, and even category of the object for cross-category texture transfer.
While previous works~\cite{bokhovkin2023mesh2tex} use CLIP~\cite{radford2021learning} similarity score, measuring it as the cosine distance between CLIP image features of the input texture image and rendered views of the generated textured mesh, we argue that it is not an appropriate metric for this task as it cannot assess the semantic awareness of texture transfer.
CLIP similarity primarily relies on matching image features, and when used to evaluate texture transfer from an image to a 3D mesh, it neglects the underlying semantics and geometry of the 3D model.
Consequently, a textured 3D model may indiscriminately mimic the input texture, leading to a high CLIP similarity score but compromise the geometric and semantic identity of the 3D mesh which is not penalized (see Fig.~\ref{fig:clip_similarity}).
As a result, we do not report the CLIP similarity score and, instead, conduct on a user study to evaluate the efficacy of our approach.

\subsection{Baseline}
\label{subsec:baseline}
We compare our approach with TEXTure~\cite{Richardson2023Texture}, which is a state-of-the-art optimization-free 3D mesh texturing technique. They also propose to transfer texture from images using personalization.
TEXTure employs a combination of Textual Inversion~\cite{gal2022textual} and DreamBooth~\cite{ruiz2023dreambooth} to first learn the target texture concept, this personalized diffusion model is then used to transfer textures to the 3D shapes.
We fine-tune TEXTure for 500 iterations, and use a single texture image, $\mathbf{I_{tex}}$, for all our experiments.

\subsection{Implementation Details} 
\label{subsec:impl_details}
Our default pipeline for transferring textures (\textit{i.e.} without \textit{Image Inversion}) does not include any network training or optimization, and we follow an entirely feed-forward process.
However, the proposed \textit{Image Inversion} involves fine-tuning parts of the network using the input texture image (see Sec.~\ref{subsubsec:img_inv}).
In contrast to baseline~\cite{Richardson2023Texture}, where the network needs to be fine-tuned for every texture image, \textit{Image Inversion} is a completely optional step, it further demonstrates the ability to learn the target concept in as few as 100 iterations when required, making it very efficient.
We fine-tune our network with a small learning rate, $1e^{-6}$, at a batch size of 4.  
We fix the value of $\lambda_{cn}$ to $1$ and found the optimal value of $\lambda_{ip}$ to vary, but to usually lie in the range $[0.2, 1.0]$, for each image-mesh pair (see Sec. \ref{subsubsec:vary_lambda_ip} for more details).
We use pre-trained Stable Diffusion v1.5~\cite{rombach2022high} as our base diffusion model, Canny ControlNet v1.1~\cite{zhang2023adding} for edge conditioning, and IP-Adapter Plus~\cite{ye2023ip} for texture image conditioning.

\subsection{Qualitative Comparison}

\input{figures/design_textransfer}

\input{figures/texture_variation}

\input{figures/opt_free_based_tex}

We show visual results comparing our approach to the baseline, TEXTure~\cite{Richardson2023Texture}, in Fig.~\ref{fig:teaser}, Fig.~\ref{fig:qualitative_comparison} and Fig~\ref{fig:clip_similarity}.
TEXTure’s failure to respect the 3D model’s geometry is evident from the cargo truck example (see Fig.~\ref{fig:clip_similarity}) where it incorrectly generates window textures on the container, highlighting its deficiency in understanding the model’s semantics.
Such a drawback lies in its reliance on depth as a conditioning cue, which lacks detailed geometric information from the mesh. To address this, we propose using edges — a more informative feature — to better align textures with the 3D model’s geometry.
In addition, we found TEXTure fails to generalize well when transferring textures between distinct object categories (such as, from a human character to a gaming controller), resulting in a loss of textural details in the generated output.
On the other hand, our method leverages IP-Adapter~\cite{ye2023ip}, and employs \textit{Image Inversion} when required, to improve fidelity to the subtle nuances of the texture image and ensure better preservation of details during the texture transfer process. 

Our approach is also capable of transferring textures from design images, as shown in Fig.~\ref{fig:design_textransfer}. We further demonstrate the ability of our approach to generate variations of texture from the same input texture image on a given 3D mesh, in Fig.~\ref{fig:texture_variation}.
Lastly, we show some results generated through optimization-based texturing (OBT)~\cite{wang2023prolificdreamer} and compare them to generations from optimization-free texturing (OFT)~\cite{Richardson2023Texture} methods in Fig.~\ref{fig:opt_free_based_tex}.
While OBT gets rid of the seams and artifacts, it takes a few hours to generate textures, whereas OFT only takes a few minutes.


\subsubsection{Comparison to Latent-NeRF}
\label{subsubsec:comp_latent_nerf}

\input{figures/latent_nerf}

While Latent-NeRF~\cite{metzer2023latent} is not designed for the problem we are solving, single-image guided 3D mesh texturing, one might still be curious about the results achievable with it. Consequently, we show some qualitative results comparing it to our approach in Fig.~\ref{fig:latent_nerf}.

\subsection{Quantitative Comparison} 

\input{tables/user_study}

We conduct a user study to evaluate and compare our approach against the baseline. In this study, 43 participants, graduate students with background in computer science, assessed the quality of generated textures with regard to both adhering to the underlying geometry of the 3D model (shape-texture consistency) and resembling the original input texture (texture fidelity).
As shown in Tab.~\ref{tab:user_study}, our approach surpasses TEXTure~\cite{Richardson2023Texture} on both the evaluation criteria, respecting the 3D model's geometry and the input texture image better at the same time.

\input{tables/time_taken}

Furthermore, TEXTure needs additional time, compute, and memory for personalization, every time it needs to transfer texture from a new image, making it difficult to use.
Tab.~\ref{tab:time_taken} shows the time taken by different methods and the time saved when compared to the baseline for learning/encoding the input texture.
We save \textasciitilde18 mins without \textit{Image Inversion}, and \textasciitilde12 mins with \textit{Image Inversion} when compared to the baseline.
However, the reported times are only for learning/encoding the input texture and does not include the time taken for texturing the 3D mesh, as one can employ any of the existing mesh texturing techniques~\cite{chen2023text2tex, Chen_2023_ICCV, Richardson2023Texture, wang2023prolificdreamer} for this process, and the time taken also varies accordingly.

\subsection{Ablation Studies}
\label{ref:ablation}
We conduct ablation experiments to examine the effect and understand the importance of various components in our pipeline.

\subsubsection{Varying $\lambda_{ip}$:}
\label{subsubsec:vary_lambda_ip}

\input{figures/vary_lambda_ip}

$\lambda_{ip}$ controls the influence of the input texture image on the diffusion output.
We gradually increase $\lambda_{ip}$ from $0$ to $1$ at an interval of $0.2$ and visualize the corresponding textured views in Fig.~\ref{fig:effect_of_lamda_ip}.
It is evident that when $\lambda_{ip}$ is small, the influence of the texture image is negligible, but as the value of $\lambda_{ip}$ increases, the input textures become more prominent in the generated output.
We believe this controllability to be an important feature that is missing in the standard personalization techniques~\cite{gal2022textual, ruiz2023dreambooth}, as these methods only allow for a fixed degree of texture transfer post-personalization. In contrast, our proposed approach enables us to control the influence of the texture image at runtime by simply varying $\lambda_{ip}$.

\subsubsection{The effect of Image Inversion:}
\label{subsubsec:effect_of_img_inv}

\input{figures/effect_of_img_inv}

To better understand the effect of \textit{Image Inversion}, we compare the results generated with pre-trained models to the ones generated employing \textit{Image Inversion}.
As shown in Fig.~\ref{fig:effect_of_image_inversion}, the results produced with \textit{Image Inversion} are significantly more detailed and faithful to the input texture than the ones from the pre-trained models. The \textit{Image Inversion} results were generated after fine-tuning the network 
for 100 iterations.

\subsubsection{Generating edge maps for conditioning:}
\label{subsubsec:gen_edge_map}

\input{figures/generating_edges}

We argue edges, as we propose to compute them, to be more detailed and serve as a stronger conditioning signal than depth or normals alone while texturing. 
We harness various geometric attributes of the input 3D mesh to extract rich edge maps for conditioning.
Fig.~\ref{fig:edge_ablation} shows the edges extracted from different geometric features (CCs, depth, and normals) of the mesh, and the corresponding textured views, separately. 
We show this for both detailed (astronaut) and simpler (cow) meshes.
While edges from CCs are the most informative for detailed high-quality meshes, they are not very helpful when it comes to simpler 3D models, instead, the edges from depth and normals are more useful in such scenarios.
Each of the edge maps capture a unique set of features, and our proposed approach combines these complementary information together to generate a final edge map, $\mathbf{I_{edge}}$, that is robust to meshes with varying levels of details, various viewpoints, etc.

%% file: figures/visual_comp.tex
\begin{figure*}[!t]
    \centering
    \includegraphics[width=0.92\textwidth, keepaspectratio]{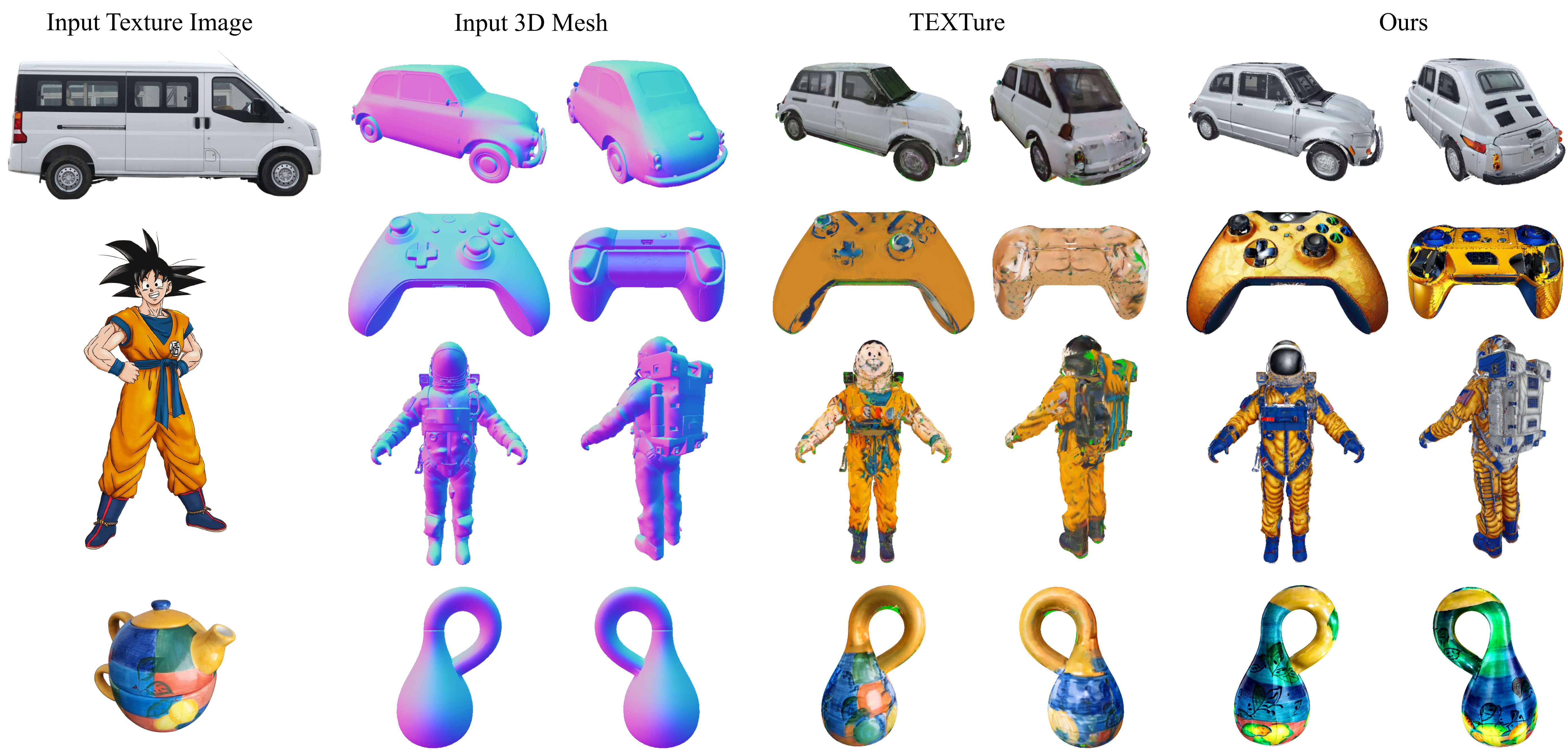}
    \caption{
    Visual comparison between results from  TEXTure~\cite{Richardson2023Texture} and our method. It is apparent that our method respects the input texture image and the underlying geometry of the 3D shape better than TEXTure.
    \textit{Image Credits:} White Van~\cite{yang2015large}, Goku~\cite{goku_img}, Teapot~\cite{gal2022textual}.
    \textit{Mesh Credits:} Car~\cite{compact_car_mesh}, Gaming Controller~\cite{xbox_mesh}, Astronaut~\cite{astronaut_high_mesh}, Klein Bottle~\cite{klein_mesh}.
    }
    \label{fig:qualitative_comparison}
\end{figure*}


%% file: figures/clip_similarity.tex
\begin{figure}
    \centering

    \includegraphics[width=\columnwidth, keepaspectratio]{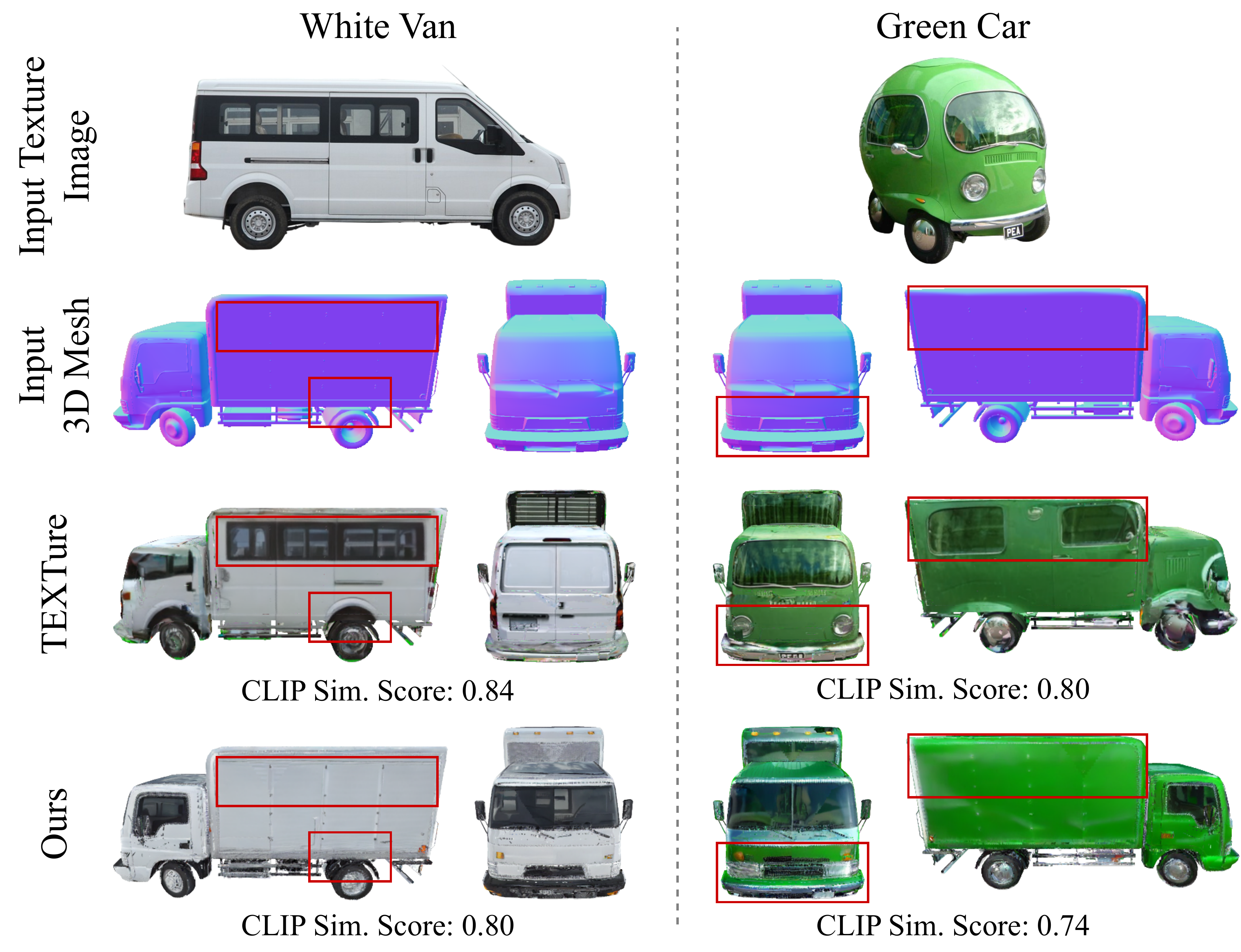}

    \caption{
    Examples showing how weak conditioning signals, such as depth, can completely alter an object’s identity during the texturing process. Specifically, we can see how a cargo truck transformed into a van when transferring textures from the white van and how the generated texture has copied the headlights and windows onto the truck when transferring textures from the green car. We also report the CLIP-similarity score (averaged across 24 views around the mesh) for all the texture transfers above. The inconsistency between CLIP-similarity scores and  generated textures for the given 3D mesh underscores the limitations of the CLIP-similarity score as a metric for evaluating the efficacy of texture transfer.
    \textit{Image Credits:} White Van~\cite{yang2015large}, Green Car~\cite{green_car_img}.
    \textit{Mesh Credits:} Cargo Truck~\cite{truck_mesh}.
    }
    \label{fig:clip_similarity}
\end{figure}

%% file: figures/design_textransfer.tex
\begin{figure}
    \centering

    \includegraphics[width=\columnwidth, keepaspectratio]{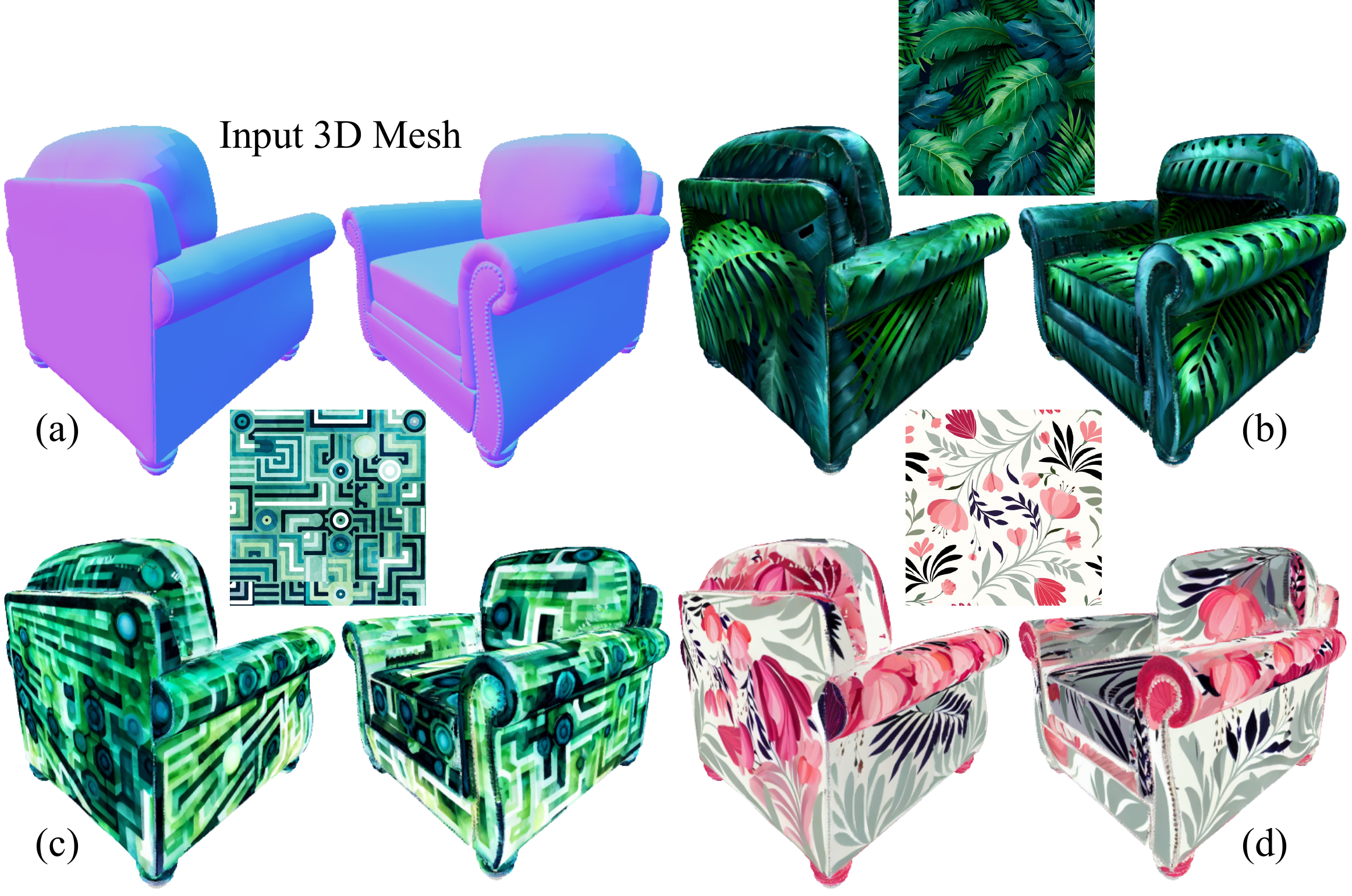}

    \caption{
    Texture transfer from design images. The input 3D mesh is shown in (a), and the generated textured meshes are shown in (b), (c), and (d). The reference texture image is shown in the middle of the textured views.
    \textit{Image Credits:} Leaves~\cite{leaves_img}, Watercolor Geometric~\cite{watercolor_geometric_img}, Floral Design~\cite{flower_pattern_img}.
    \textit{Mesh Credits:} Single Sofa~\cite{sofa_mesh}.
    }
    \label{fig:design_textransfer}
\end{figure}

%% file: figures/texture_variation.tex
\begin{figure}
    \centering

    \includegraphics[width=\columnwidth, keepaspectratio]{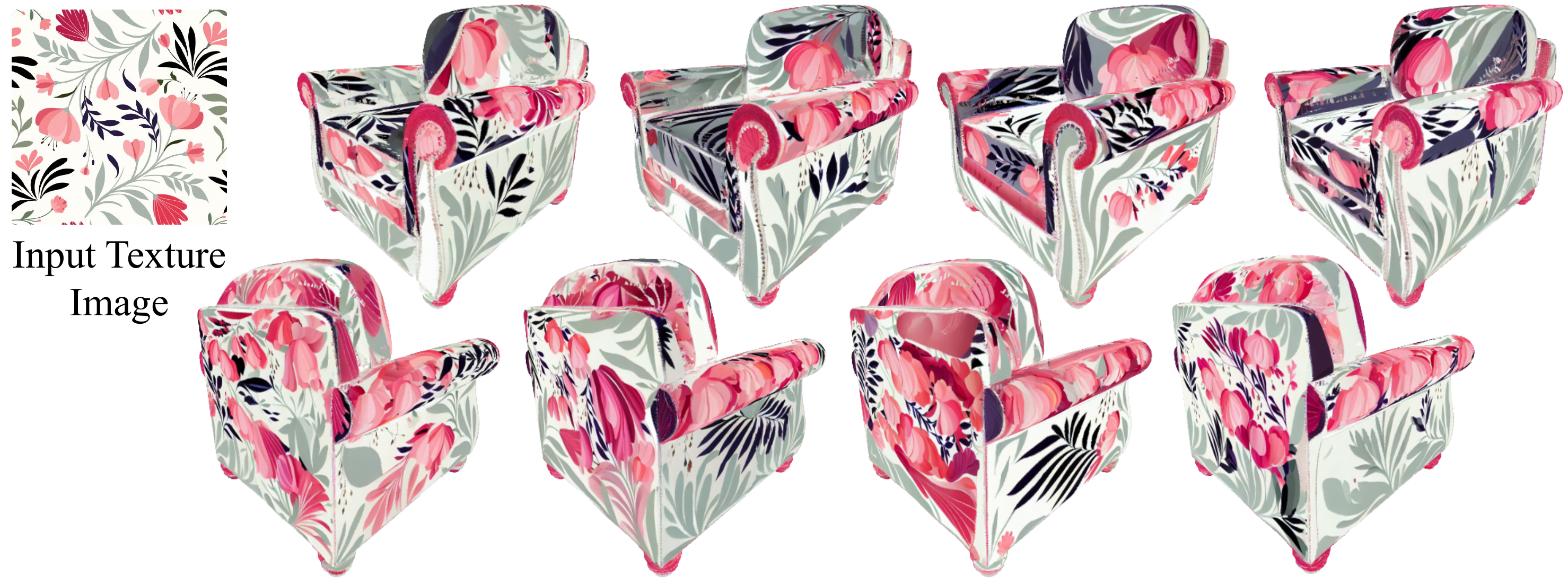}

    \caption{
    Texture Variation: The diversity of textures synthesized from a single texture image on the same mesh. Each column corresponds to a distinct texture generated for the mesh by varying the seed.
    \textit{Image Credits:} Floral Design~\cite{flower_pattern_img}.
    \textit{Mesh Credits:} Single Sofa~\cite{sofa_mesh}.
    }
    \label{fig:texture_variation}
\end{figure}

%% file: figures/opt_free_based_tex.tex
\begin{figure}
    \centering

    \includegraphics[width=0.8\columnwidth, keepaspectratio]{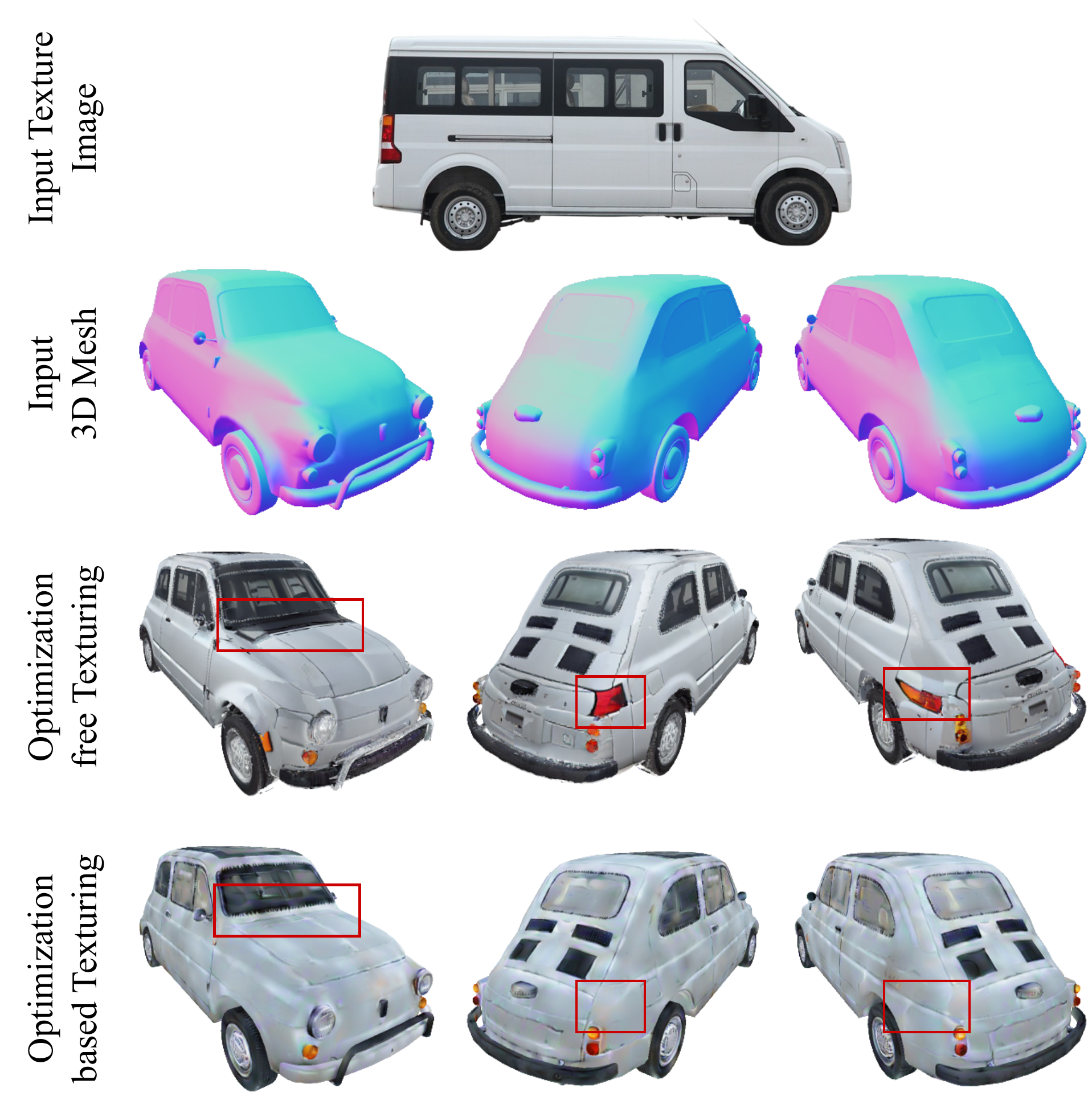}

    \caption{
    A visual comparison of results generated with optimization-free~\cite{Richardson2023Texture} and optimization-based~\cite{wang2023prolificdreamer} texturing methods, while transferring textures using our proposed approach.
    \textit{Image Credits:} White Van~\cite{yang2015large}.
    \textit{Mesh Credits:} Car~\cite{compact_car_mesh}.    
    }
    \label{fig:opt_free_based_tex}
\end{figure}

%% file: figures/latent_nerf.tex
\begin{figure}
    \centering

    \includegraphics[width=\columnwidth, keepaspectratio]{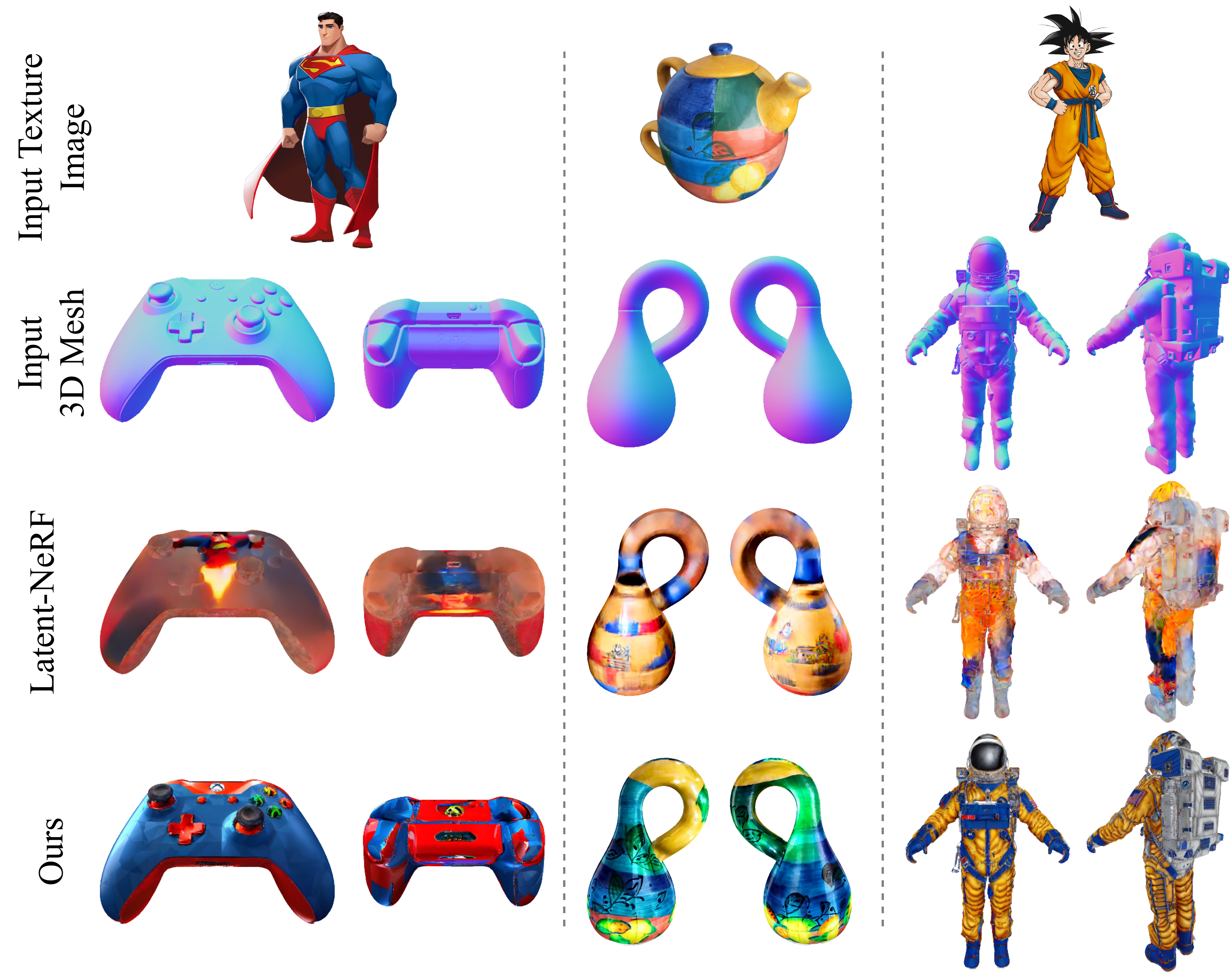}

    \caption{
    A visual comparison of results from Latent-NeRF~\cite{metzer2023latent} to ours.
    \textit{Image Credits:} Superman~\cite{superman_img}, Teapot~\cite{gal2022textual}, Goku~\cite{goku_img}.
    \textit{Mesh Credits:} Gaming Controller~\cite{compact_car_mesh}, Klein Bottle~\cite{klein_mesh}, Astronaut~\cite{astronaut_high_mesh}.
    }
    \label{fig:latent_nerf}
\end{figure}

%% file: tables/user_study.tex
\begin{table}
    \centering
    \caption{
        User study showing the percentage (\%) of people preferring results from an approach for different evaluation criteria. Our approach outperforms TEXTure~\cite{Richardson2023Texture}, both in terms of shape-texture consistency and texture fidelity, indicating that it respects the input 3D shape and texture image simultaneously. 
    }
    \label{tab:user_study}
    \begin{tabular}{ccc}
        \toprule
        Evaluation Criteria       & TEXTure & Ours \\ 
                                  \midrule
        Shape-Texture Consistency &   19.53      &   \textbf{80.47}   \\
        Texture Fidelity     &   29.53      &   \textbf{70.47}  \\ 
        \bottomrule
    \end{tabular}
\end{table}


%% file: tables/time_taken.tex
\begin{table}
    \centering
    \caption{
    Table showing the time taken by different methods to learn/encode the input texture image and the time saved when compared to TEXTure~\cite{Richardson2023Texture}, in the last column.
    We compare TEXTure to our approach, both with and without \textit{Image Inversion (Img. Inv.)}.
    }
    \label{tab:time_taken}
    \begin{tabular}{ccc}
    \toprule
            Method & Time taken & Time saved \\ 
            \midrule
            TEXTure &   \textasciitilde18 mins      &   -  \\
            Ours (w/ \textit{Img. Inv.}) &   \textasciitilde6 mins      &   \textasciitilde12 mins (66\%)  \\
            Ours (w/o \textit{Img. Inv.}) &   <100 ms      &   \textasciitilde18 mins (100\%)  \\
    \bottomrule
    \end{tabular}
\end{table}


%% file: figures/vary_lambda_ip.tex
\begin{figure}
    \centering

    \includegraphics[width=\columnwidth, keepaspectratio]{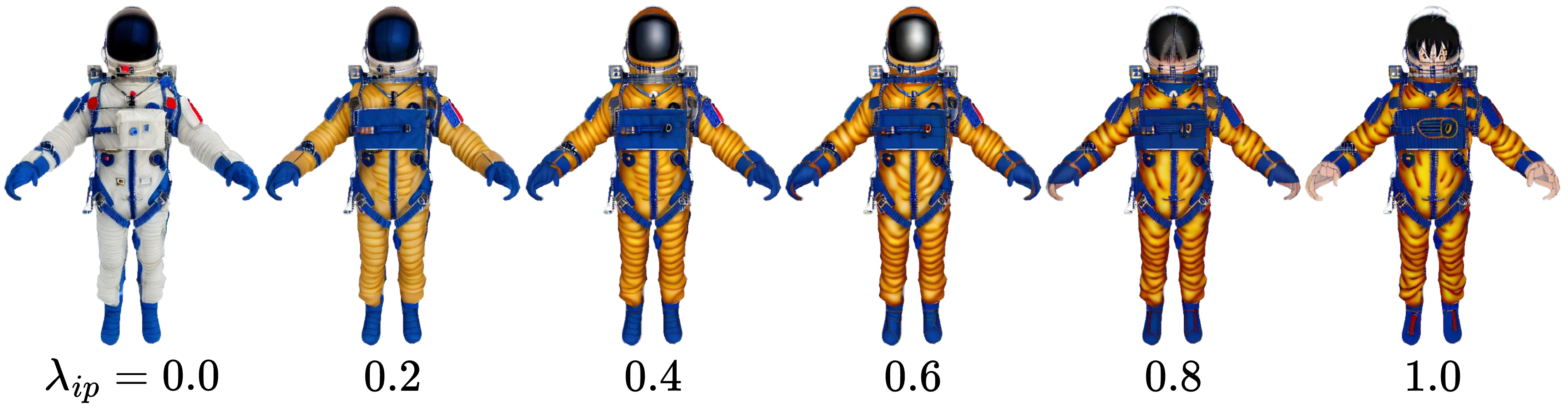}

    \caption{
    The effect of varying $\lambda_{ip}$ on the generated output. The input 3D mesh and texture image are the same as in Fig.~\ref{fig:depth_vs_edge}.
    }
    \label{fig:effect_of_lamda_ip}
\end{figure}

%% file: figures/effect_of_img_inv.tex
\begin{figure}
    \centering

    \includegraphics[width=\columnwidth, keepaspectratio]{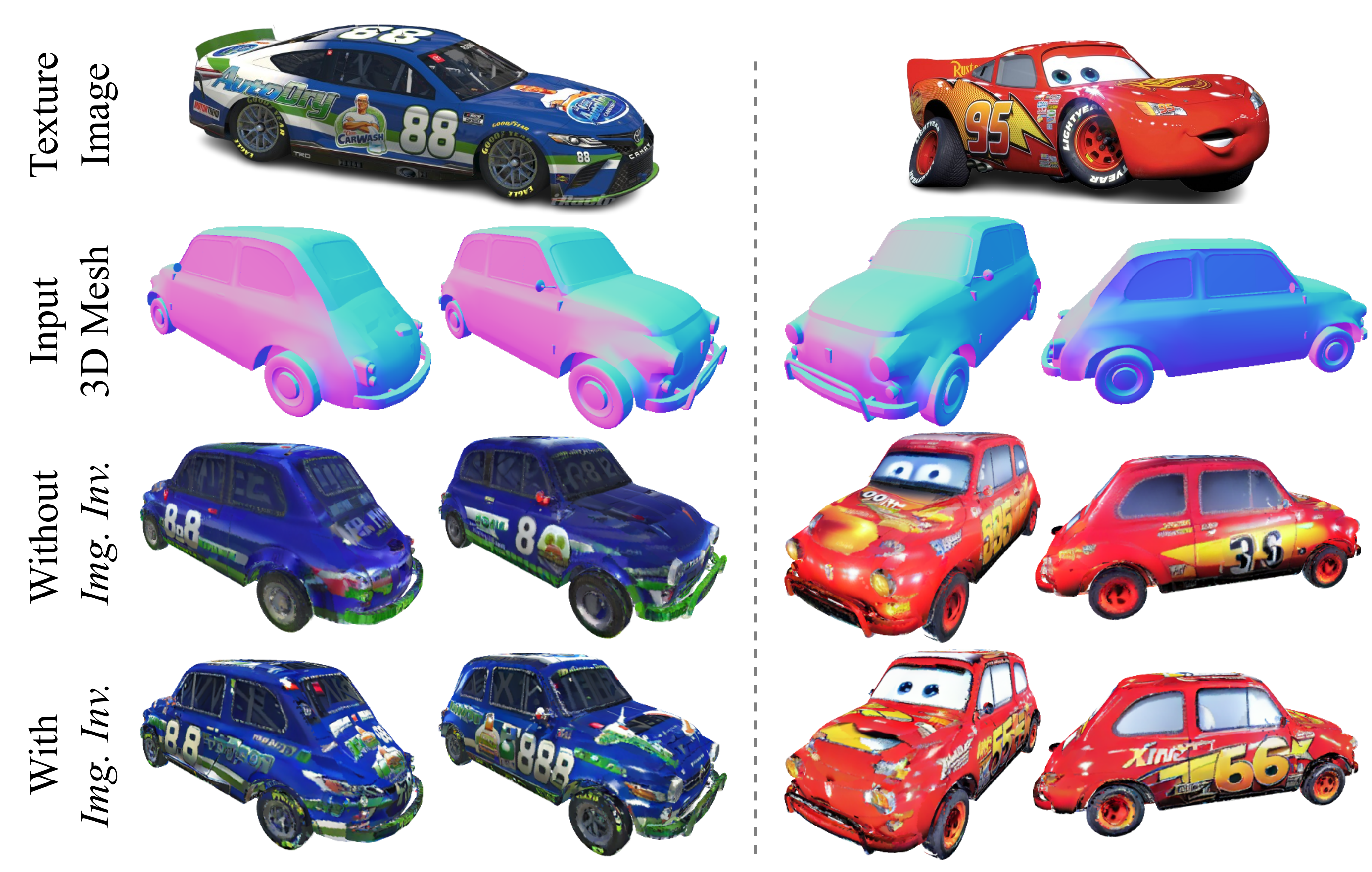}

    \caption{
    Results generated from our approach, with and without \textit{Image Inversion (Img. Inv.)}.
    \textit{Image Credits:} Nascar~\cite{nascar_img}, Lightning McQueen~\cite{lmq_img}.
    \textit{Mesh Credits:} Car~\cite{compact_car_mesh}.
    }
    \label{fig:effect_of_image_inversion}
\end{figure}

%% file: figures/generating_edges.tex
\begin{figure}
    \centering

    \includegraphics[width=0.8\columnwidth, keepaspectratio]{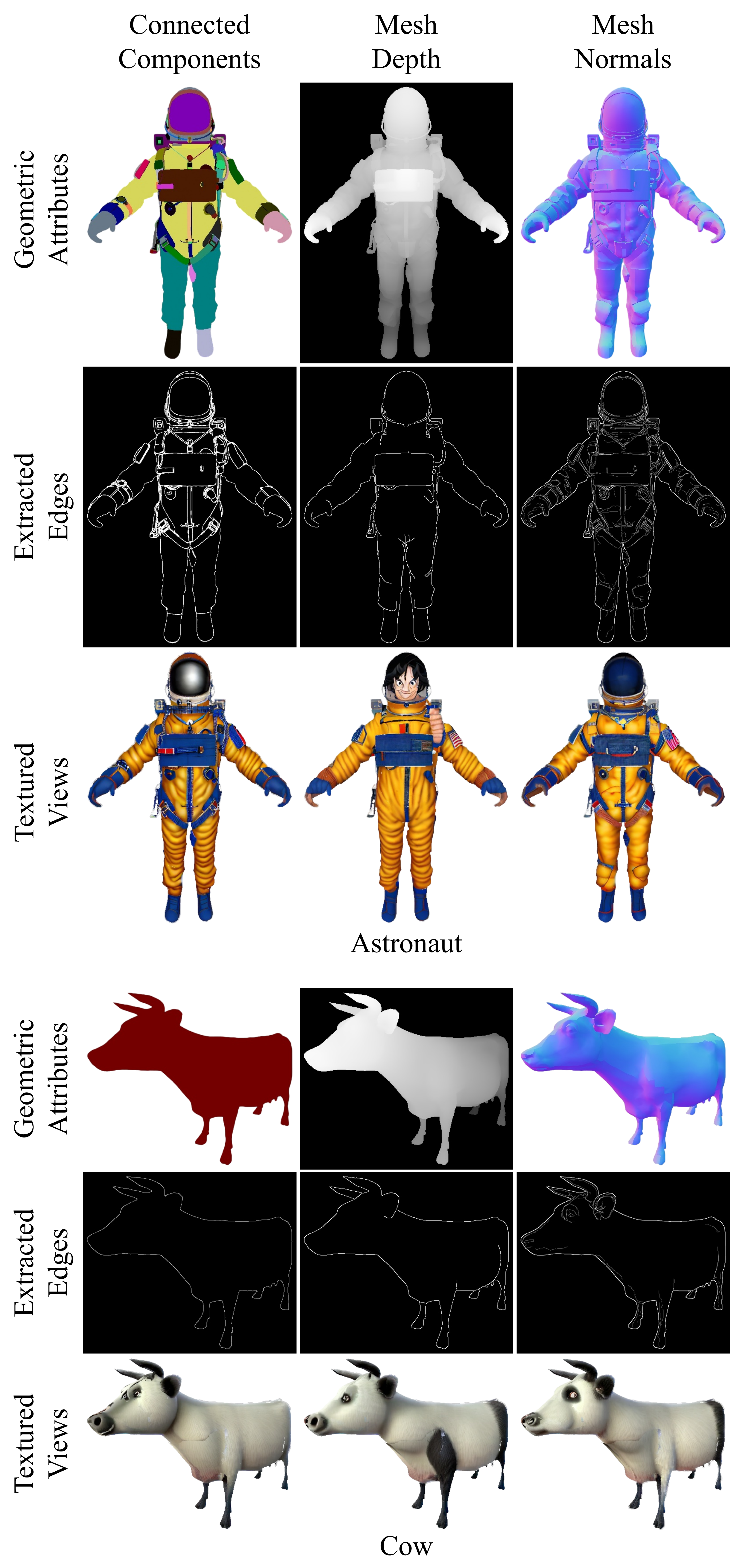}

    \caption{
    The edges extracted from different geometric attributes (CCs, depth, and normals) of the mesh, and the corresponding textured views. We show this for both detailed (astronaut) and simpler (cow) meshes. Zoom in to see the edge maps in detail. 
    The input 3D meshes and texture images are the same as in Fig.~\ref{fig:teaser} and Fig.~\ref{fig:depth_vs_edge}.
    }
    \label{fig:edge_ablation}
\end{figure}

%% file: sections/5_conclusion.tex
\section{Conclusion}
\label{sec:conclusion}


We propose EASI-Tex, a novel approach for single-image guided edge-aware 3D mesh texturing. We treat this as a generative task and employ a pre-trained Stable Diffusion~\cite{rombach2022high} model with judicious conditioning to achieve high-quality texture transfers.
To preserve the \textit{identity} of the mesh while texturing, we condition the diffusion model with edges extracted from various geometric features (CCs, normals, and depth) of the mesh using ControlNet~\cite{zhang2023adding}.
We further condition the diffusion model with the reference texture image using IP-Adapter~\cite{ye2023ip}, which allows us to use a single image as prompt without any additional training or optimization. 
We also introduce \textit{Image Inversion}, a new personalization technique to quickly adapt the diffusion model for a single concept using a single image, for cases where the pre-trained IP-Adapter falls short in capturing all the details from the input image faithfully.
Experimental results show the superiority of our approach in generating high-quality textures that respect the input texture image and the underlying geometry of the 3D mesh, in an efficient manner.
We also demonstrate the robustness of our approach in handling meshes with varying levels of detail, as well as its ability to tackle challenging image-mesh pairs through experiments.


The main limitation of our approach in terms of transferring textures is the input resolution of the CLIP image encoder in the IP-Adapter.
While CLIP serves as a robust image encoder, its input is restricted to $224 \times 224$ images. Such a constraint hinders its ability to capture small intricate details
from the input texture image faithfully and subsequently transfer them to the 3D shape while texturing.
Although our proposed \textit{Image Inversion} mitigates this issue to a certain extent, an exciting direction for further exploration lies in modifying the IP-Adapter to accommodate larger image inputs.

In addition, texture seams may occasionally appear in our generated textures. These stem from the mesh texturing technique, Text2Tex~\cite{chen2023text2tex}, which employs an iterative texture pasting strategy. These seams can be avoided by using optimization-based techniques~\cite{Chen_2023_ICCV, wang2023prolificdreamer} for texturing (see Fig.~\ref{fig:opt_free_based_tex}). However, there is a trade-off: as outlined in Sec.~\ref{subsec:tex_mesh}, these methods require significantly more time, often hours, compared to the optimization-free approach we took, which usually takes minutes.

%% file: sections/6_acknowledgement.tex

We thank all the anonymous reviewers for their insightful comments and constructive feedback. Thanks also go to Aditya Vora and Elad Richardson for helpful discussions. This work was supported in part by NSERC, gift funds from Adobe, and AWS Credits from Amazon.